\documentclass[sigconf]{acmart}

\usepackage{algorithm}
\usepackage{algorithmic}
\usepackage{amsmath}
\usepackage{multirow}
\usepackage{array}
\usepackage{booktabs} 
\usepackage{xcolor}
\usepackage[inline]{enumitem}

\usepackage{amsthm}
\usepackage{makecell}

\newtheoremstyle{noindentstyle}
  {}     
  {}     
  {\normalfont} 
  {0pt}  
  {\bfseries} 
  {.}    
  { }    
  {}     

\theoremstyle{noindentstyle}

\newlist{desclist}{description}{3}
\setlist[desclist,1]{format=\preitem\bfseries,leftmargin=\widthof{\preitem},style=sameline}

\newcommand\preitem{\mdseries\textbullet\space}
\newcommand{\name}{{\textsc{EviNet}}}

\newtheorem{problem}{Problem}

\AtBeginDocument{%
  }

\copyrightyear{2025}
\acmYear{2025}
\setcopyright{cc}
\setcctype{by}
\acmConference[KDD '25]{Proceedings of the 31st ACM SIGKDD Conference on
Knowledge Discovery and Data Mining V.2}{August 3--7, 2025}{Toronto, ON,
Canada}
\acmBooktitle{Proceedings of the 31st ACM SIGKDD Conference on Knowledge
Discovery and Data Mining V.2 (KDD '25), August 3--7, 2025, Toronto, ON,
Canada}
\acmDOI{10.1145/3711896.3736945}
\acmISBN{979-8-4007-1454-2/2025/08}

\begin{document}

\title{\textsc{\name:}Towards Open-World Graph Learning via Evidential Reasoning Network}

\author{Weijie Guan}
\email{skjguan@vt.edu}
\affiliation{%
  \institution{Virginia Polytechnic Institute and State University}
  \city{Blacksburg}
  \state{VA}
  \country{USA}
}

\author{Haohui Wang}
\email{haohuiw@vt.edu}
\affiliation{%
  \institution{Virginia Polytechnic Institute and State University}
  \city{Blacksburg}
  \state{VA}
  \country{USA}
}

\author{Jian Kang}
\email{jian.kang@rochester.edu}
\affiliation{%
  \institution{University of Rochester}
  \city{Rochester}
  \state{NY}
  \country{USA}
}

\author{Lihui Liu}
\email{hw6926@wayne.edu}
\affiliation{%
  \institution{Wayne State University}
  \city{Detroit}
  \state{MI}
  \country{USA}
}

\author{Dawei Zhou}
\email{zhoud@vt.edu}
\affiliation{%
  \institution{Virginia Polytechnic Institute and State University}
  \city{Blacksburg}
  \state{VA}
  \country{USA}
}

\renewcommand{\shortauthors}{Weijie Guan, Haohui Wang, Jian Kang, Lihui Liu, \& Dawei Zhou}

\begin{abstract}
Graph learning has been crucial to many real-world tasks, but they are often studied with a closed-world assumption, with all possible labels of data known a priori. To enable effective graph learning in an open and noisy environment, it is critical to inform the model users when the model makes a wrong prediction to in-distribution data of a known class, i.e., misclassification detection or when the model encounters out-of-distribution from novel classes, i.e., out-of-distribution detection. This paper introduces Evidential Reasoning Network (\name), a framework that addresses these two challenges by integrating Beta embedding within a subjective logic framework. \name~includes two key modules: Dissonance Reasoning for misclassification detection and Vacuity Reasoning for out-of-distribution detection. Extensive experiments demonstrate that \name~outperforms state-of-the-art methods across multiple metrics in the tasks of in-distribution classification, misclassification detection, and out-of-distribution detection. \name~demonstrates the necessity of uncertainty estimation and logical reasoning for misclassification detection and out-of-distribution detection and paves the way for open-world graph learning. Our code and data are available at \url{https://github.com/SSSKJ/EviNET}.
\end{abstract}

\ccsdesc[500]{Computing methodologies~Probabilistic reasoning}

\keywords{Out-of-distribution detection, open-world machine learning}


\maketitle

\section{Introduction}
\begin{figure}[t]
\centering
\includegraphics[width=1\columnwidth]{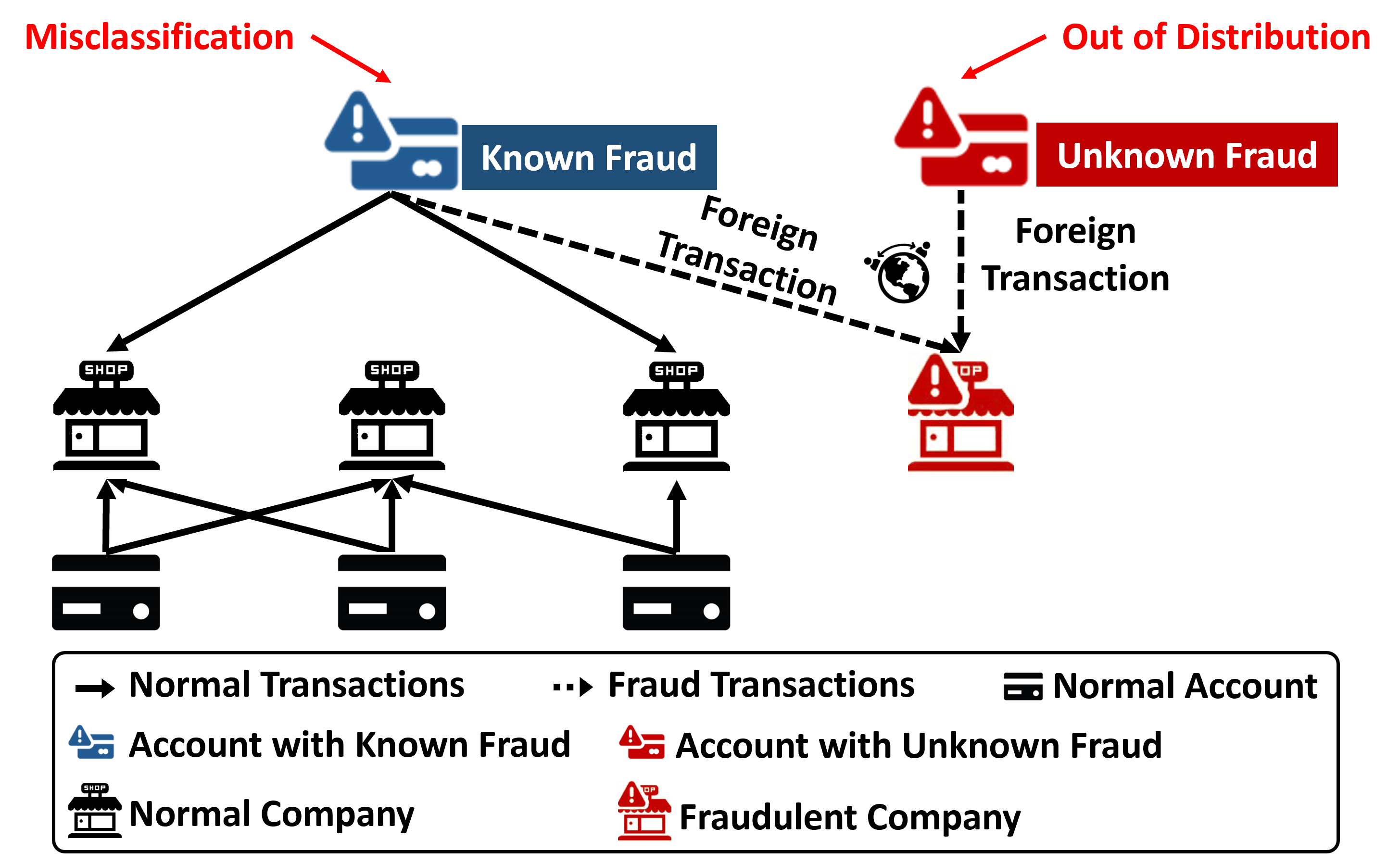} 
\caption{An illustrative figure of financial fraud situations in real-world applications. There are many undiscovered types of fraud, and even known fraudulent activities may incorporate legitimate transactions to mimic normal users and conduct noisy and covert fraud transactions, making detection increasingly challenging.}
\label{fig:financial fraud}
\end{figure}

Graphs capture relationships among entities in many real-world scenarios, including social networks~\cite{khanam2023homophily, DBLP:conf/kdd/WangJDZCZF0024}, recommendation systems~\cite{10.1145/3535101, wu2025bridging}, and financial fraud detection~\cite{wang2022review, tuo2025}. To date, researchers have developed a wealth of graph learning models, such as graph neural networks, to learn node representations on graphs~\cite{kipf2017semisupervised, NIPS2017_5dd9db5e, v2018graph}. These methods often learn the representation of a node by aggregating information from its neighbors and updating the model parameters by its label. An underlying assumption for training these graph learning models is that all possible labels are known a priori in a closed label set during training, which we term learning in \textit{a closed-world environment}.

Yet, the assumption about the closed-world environment might often be violated in real-world applications due to ambiguous data or unknown and novel classes. Graph learning models trained with the closed-world assumption might fail in high-stakes applications such as financial fraud detection~\cite{haohui2025EvoluNet, DBLP:conf/nips/WangGCW024}. 
For example, nearly 60\% of banks, fintechs, and credit unions faced over \$500K in direct fraud losses in 2023. 
Though financial fraud detection has been actively studied for years and has shown strong performance in detecting known and commonly happened types of fraud,
many types of fraud remain undiscovered (as the account marked as out-of-distribution shown in Figure \ref{fig:financial fraud}). 
Meanwhile, there could also be accounts that mostly 
engage in normal transactions but occasionally reveal fraudulent intentions (as the account marked as misclassification shown in Figure~\ref{fig:financial fraud}), particularly in cross-border money laundering 
where 
different 
regulations across countries create noise in the data 
that might mislead the model to misclassify data.
Thus, it is critical to operationalize a graph learning model in \textit{open and noisy environments} by informing the model users (\textit{C1. In-distribution misclassification detection}) when the graph learning model misclassifies data into one of the known classes or (\textit{C2. Out-of-distribution detection}) when it encounters out-of-distribution (OOD) data from novel classes that do not appear during training. 

Recognizing the importance, researchers have developed a collection of models for misclassification detection~\cite{moon2020confidenceawarelearningdeepneural, NEURIPS2021_2cb6b103} or OOD detection~\cite{liang2020enhancingreliabilityoutofdistributionimage, hendrycks2022scalingoutofdistributiondetectionrealworld, wu2023energybasedoutofdistributiondetectiongraph}. However, these existing techniques are often limited to tackling one facet of challenges in learning in open and noisy environments (i.e., either misclassification detection or OOD detection) and fail to address the multifaceted challenges posed by open and noisy environments. A few models, such as GPN~\cite{stadler2021graphposteriornetworkbayesian} and GKDE~\cite{NEURIPS2020_968c9b4f}, are capable of detecting both misclassification and OOD instances. However, GPN~\cite{stadler2021graphposteriornetworkbayesian} relies heavily on the approximate posterior distributions of known classes to make predictions, which might make it difficult to detect subtle misclassifications. Moreover, GPN quantifies uncertainty based on the posterior distribution of known classes. It might not generate meaningful representations for detecting novel classes that are out-of-distribution.
Regarding GKDE~\cite{NEURIPS2020_968c9b4f}, it is a subjective-logic-based framework with a fixed prior weight that is equal to the number of known classes and with a variance-minimization regularization that minimizes the evidence with respect to a class if a data point does not belong to that class. This essentially leads to overconfidence in both misclassification and out-of-distribution detection~\cite{chen2024redl}. Thus, it is critical to understand how to achieve effective misclassification detection and out-of-distribution detection while maintaining strong performance when classifying in-distribution data.

To this end, we propose a subjective-logic-guided framework named Evidential Reasoning Network (\name). \name~learns node embeddings using the Beta embedding to enable logical reasoning capability and provides support regions for known classes and an implicit support region for novel classes. Based on that, \name~quantifies uncertainty for misclassification detection and out-of-distribution detection with two key modules: (M1) Dissonance Reasoning that detects misclassified data and (M2) Vacuity Reasoning to detect out-of-distribution data from novel classes. Specifically, (M1) Dissonance Reasoning employs logical reasoning to generate class embeddings for known classes and computes a dissonance score using both class embeddings and node embeddings. A high dissonance score refers to a node that has a high probability of being classified into more than one class, which indicates potential misclassification. (M2) Vacuity Reasoning considers the potential novel class and derives representations for potential novel classes by logical disjunction and negation of the known class embeddings in Dissonance Reasoning. Then, it detects out-of-distribution data from novel classes by computing a vacuity score using node embeddings, class embeddings of known classes, and class embedding for novel classes. By combining these two modules, \name~addresses the challenge of overconfidence and provides more reliable uncertainty estimates for both misclassification and OOD detection. We show that \name~demonstrates strong performance in both misclassification and out-of-distribution detection while maintaining competitive performance for in-distribution classification. Our main contributions are as follows:

\begin{itemize}[leftmargin=*]
    \item \textbf{Problem.} We study graph learning in an open and noisy environment by addressing misclassification detection and OOD detection simultaneously.
    \item \textbf{Algorithm.} We propose 
    \name~which employs Beta embedding and subjective logic
    , providing support regions for known classes and an implicit support region for novel classes to estimate uncertainty for both misclassification and OOD detection. 
    \item \textbf{Evaluation.} We systematically evaluate the performance of \name~in in-distribution classification, misclassification detection, and OOD detection. Extensive experiments show the efficacy of \name~in all three tasks.
\end{itemize}

The rest of this paper is organized as follows. We provide background information and problem definition in Section 2, followed by the proposed framework in Section 3. In Section 4, we discuss the experimental setup and results, followed by a literature review in Section 5. Finally, we conclude the paper in Section 6.

\section{Preliminary}
In this section, we first present key symbols and notations in this work and then offer a preliminary review of subjective logic and Beta embedding, which is core to our proposed \name.

\noindent{\textbf{Notations.}} In general, we represent a vector, a matrix, and a set with a bold lowercase letter ($\boldsymbol{b}$), bold uppercase letter ($\mathbf{X}$), and calligraphic letter ($\mathcal{V}$). In this work, we denote a graph as $\left(\mathbf{A}, \mathbf{X}, \mathcal{Y}\right)$ or $\left(\mathbf{A}, \mathbf{X}\right)$ without labels, where $\mathbf{A}$ is its adjacency matrix determined by a set of $n$ nodes $\mathcal{V}$ and a set of $m$ edges $\mathcal{E}$, $\mathbf{X}$ is the node feature matrix, and $\mathcal{Y}$ is the label set. Table~\ref{tab:symbols} in Appendix~\ref{sec:notation} summarizes key symbols and their descriptions.

\noindent{\textbf{Subjective Logic.}}
It is a probabilistic logic framework to manage belief and uncertainty and make inferences under incomplete and uncertain information \cite{josang2018uncertainty}. Subjective opinions in subjective logic quantifies the beliefs about the truth of propositions under uncertainty, which could be a binomial opinion, multinomial opinion, or hyper opinion. We consider multinomial opinion because it considers beliefs for multiple possible outcomes and matches with the multi-class classification problem in our setting. Specifically, for a multi-class classification with label set $\mathcal{Y} = \{1, \ldots, K\}$, a multinomial opinion over $\mathcal{Y}$ is a triplet $\omega = \left(\boldsymbol{b}, u, \boldsymbol{a}\right)$, where $\boldsymbol{b} = \left[b_1, \ldots, b_K\right]^T$ is the belief mass distribution over $\mathcal{Y}$, $u$ is the uncertainty mass that represents vacuity of evidence, and $\boldsymbol{a} = \left[a_1, \ldots, a_K\right]^T$ is the base rate distribution over $\mathcal{Y}$. The multinomial opinion can be further represented as a Dirichlet distribution of the projected probability distribution $\boldsymbol{p} = \left[p_1, \ldots, p_K\right]$.
Specifically, the expected probability for class $k \in \mathcal{Y}$ is
\begin{equation}
    \text{E}\left[p_k\right] = \frac{\xi_k}{\sum_{i=1}^K \xi_i} = \frac{e_k + a_k W}{W + \sum_{i=1}^K e_i},
\end{equation}
where $\boldsymbol{\xi} = \{\xi_1, \ldots, \xi_K\}$ is a vector of evidence parameters, $e_k$ is the evidence for class $k$, and $W$ is the prior weight representing uncertain evidence. To map the observed evidence to a multinomial opinion, the belief and uncertainty are computed as
\begin{equation}
    b_k = \frac{e_k}{\sum_{i=1}^K \xi_i} = \frac{e_k}{W + \sum_{i=1}^K e_i}, \quad u = \frac{W}{\sum_{i=1}^K \xi_i} = \frac{W}{W + \sum_{i=1}^K e_i},
\end{equation}
such that the projected probability for class $k$ is $p_k = b_k + a_k u$.
Given the multinomial opinion $\omega = \left(\boldsymbol{b}, u, \boldsymbol{a}\right)$, we can further understand the belief vacuity resulted from lack of knowledge as 
\begin{equation}\label{eq:vacuity}
    \operatorname{Vac}\left(\omega\right) = u = \frac{W}{W + \sum_{i=1}^K e_i}
\end{equation}
and the dissonance, which reflects conflicting evidence, as 
\begin{equation}\label{eq:dissonance}
    \operatorname{Diss}\left(\omega\right) = \sum_{i = 1}^{K} \left(\frac{b_i \sum_{j \neq i} b_j \operatorname{Bal}\left(b_j, b_i\right)}{\sum_{j \neq i} b_j}\right)
\end{equation}
where $\operatorname{Bal}\left(b_j, b_i\right) = 1 - \left|b_j - b_i\right| /\left(b_j + b_i\right)$.

\noindent{\textbf{Beta Embedding.}} It is originally designed for logical reasoning on knowledge graphs~\cite{ren2020beta}. Beta embedding is a probabilistic embedding framework to model uncertainty and perform logical operations within the embedding space. Specifically, a Beta embedding in a $d$-dimensional space is represented by a collection of independent Beta distributions $\left\{\left(\alpha_{1}, \beta_{1}\right), \ldots, \left(\alpha_{d}, \beta_{d}\right)\right\}$, where each pair of parameters $(\alpha, \beta)$ defines a Beta distribution. Then neural logical operators such as projection, intersection, and negation are applied by equipping these Beta distributions with neural networks.

\noindent{\textbf{Problem Definition.}} Graph Learning in Open and Noisy Environments is an important problem in open environments. A reliable model should not only output the classification prediction for samples but also assess how likely it is that a given sample is misclassified or belongs to novel classes. Most existing methods focus only on OOD detection \cite{openfacesurvey, 9163102, Gunther_2017_CVPR_Workshops, Zhao_2023_CVPR} or misclassification detection \cite{NEURIPS2021_2cb6b103, vazhentsev-etal-2022-uncertainty}. This limited focus reduces the model’s overall effectiveness and robustness in real-world applications. Building a reliable model for real-world environments requires a comprehensive approach that can jointly address both OOD detection and misclassification detection, ensuring that the model can handle the uncertainties inherent in these environments. Based on the above considerations, the problem addressed in this paper is formally defined as follows:

\begin{problem}{Graph learning in an open and noisy environment.}
    \\
    \textbf{Given:} An undirected attributed graph $\left(\mathbf{A}, \mathbf{X}, \mathcal{Y}\right)$. 
    \textbf{Find:} A model $f\left(\cdot\right)$ that, for each node, outputs (1) the predicted label, (2) a misclassification score indicating the likelihood that the prediction is misclassified, and (3) an out-of-distribution score indicating the likelihood that the node is out-of-distribution.

\end{problem}

\section{Proposed Method}

In this section, we introduce \name, a unified framework to address both (\textit{C1}) in-distribution misclassification detection and (\textit{C2}) OOD detection. The core idea of EviNet is to estimate uncertainty by representing nodes and classes as Beta embeddings and reasoning with subjective logic. 
In particular, we first introduce the overall learning framework of \name. Then, we discuss two key modules in \name, including (M1) Dissonance Reasoning for misclassification detection and (M2) Vacuity Reasoning for out-of-distribution detection. Finally, we present an end-to-end optimization framework for training \name.

\subsection{Framework Overview}
\begin{figure*}[t]
\centering
\includegraphics[width=1\textwidth]{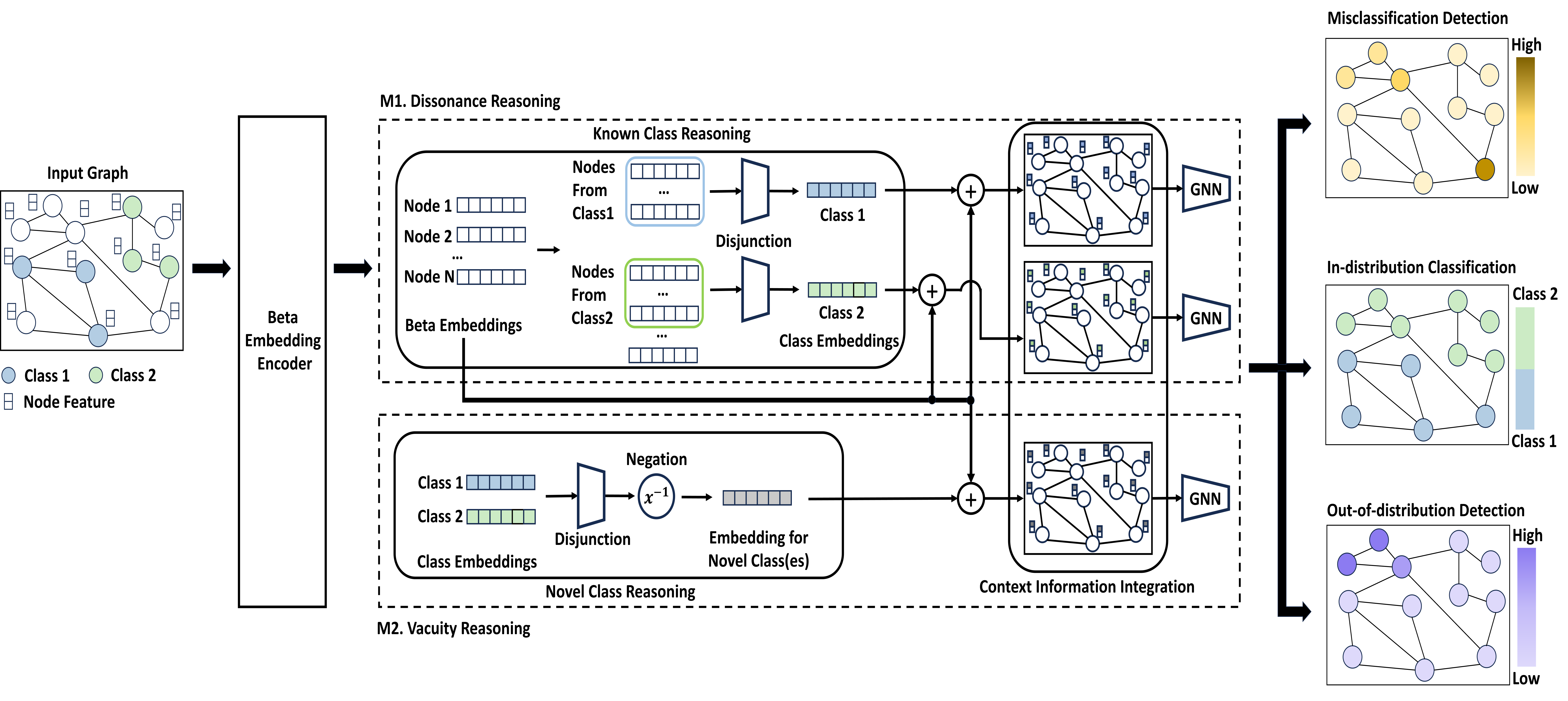} 
\caption{An overview of \name. The framework processes graph-structured data through two core modules: (1) Dissonance Reasoning (M1); (2) Vacuity Reasoning (M2). The final outputs include in-distribution classification, misclassification detection scores, and OOD detection scores. For simplicity, we consider a binary classification scenario in this example, in which Class 1 is shown in blue, Class 2 is shown in green, and Novel Class(es) is shown in grey.}
\label{fig:framework}
\end{figure*}

An overview of our proposed framework is presented in Figure~\ref{fig:framework}, which consists of two key modules: \textbf{(M1) Dissonance Reasoning} and \textbf{(M2) Vacuity Reasoning}. The Dissonance Reasoning module tackles (\textit{C1}) in-distribution misclassification detection by generating class embeddings for known (in-distribution) classes from nodes and computing a dissonance score, which indicates the conflicting evidence towards multiple known classes, using the node embeddings and class embeddings.

The Vacuity Reasoning module builds on the generated class embeddings to address (\textit{C2}) OOD detection. The core idea is to infer a class embedding for the novel (OOD) classes via logical disjunction and negation and compute the vacuity score, which indicates the lack of evidence with respect to the novel classes, using the node embeddings and the class embeddings for known classes and novel classes. 

What is central to \name~is the node representations that maintain structural and attributive information and are appropriate for logical reasoning. Classic node representation learning approaches~\cite{kipf2016variationalgraphautoencoders, grover2016node2vecscalablefeaturelearning} often map a node to a data point in a low-dimensional space that could reconstruct the graph. Though widely used, these approaches cannot effectively represent support regions of classes and reason under uncertainty. To address these limitations, in \name, we adopt Beta embedding~\cite{ren2020beta} that represents each node as a set of Beta distributions and conceptualizes a node as a set containing itself only. A prominent advantage of Beta embedding is its closure property under various logical operations, such as disjunction and negation, which are two core logical operations to reason for both known and novel classes.

\begin{figure}[h]
\centering
\includegraphics[width=0.9\columnwidth]{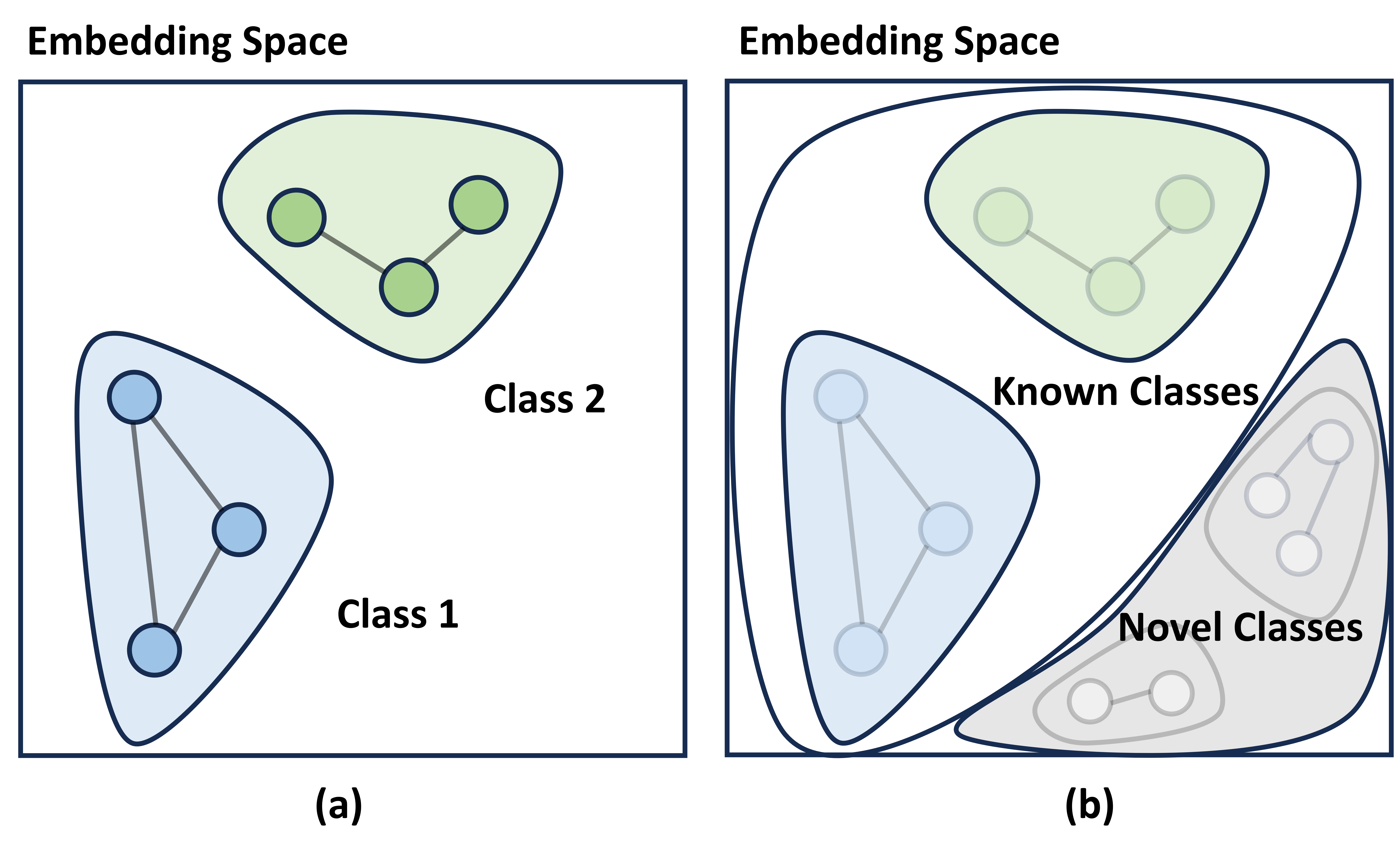}
\caption{An illustrative figure for M1. Dissonance Reasoning (a) and M2. Vacuity Reasoning (b) with two known classes. In (a), the support regions of class 1 and class 2 are generated by applying disjunction on training samples. In (b), with the support regions of class 1 and class 2, the support region involving all known classes can also be generated with disjunction and by applying negation on it, the inexplicit support region of including novel classes can be derived.}
\label{fig:modules}
\end{figure}

\subsection{M1. Detecting Misclassification via Dissonance Reasoning}\label{subsec:dissonance-reasoning}
To detect misclassification, it is important to understand to what extent the model is uncertain in its predictions. \name~builds upon subjective logic to quantify uncertainty~\cite{NEURIPS2020_968c9b4f}, because dissonance in subjective logic reflects the contradicting beliefs about a model prediction. However, the subjective logic-based model may suffer from overconfidence~\cite{chen2024redl}. One possible reason is the lack of consideration for context information between samples and classes in traditional subjective logic-based approaches.

To address this limitation, the key idea of the Dissonance Reasoning module is two-fold: First, we encode nodes as Beta embeddings and reason for the class embeddings using logical disjunction. Second, we generate class-specific graphs to produce context-aware node embeddings, which are further used to estimate dissonance uncertainty.

More specifically, let $f(\mathbf{A}, \mathbf{x}|\mathbf{\theta})$ represent an encoder that outputs a Beta embedding for the node in graph $\mathbf{A}$ with feature $\mathbf{x}$ with $\theta$ being the parameters of the encoder.
Then a $d$-dimensional Beta embedding of a given node $i$ can be presented as $\mathcal{N}_i = (\boldsymbol{\alpha}_i, \boldsymbol{\beta}_i) = \left\{\left(\alpha_{i1}, \beta_{i1}\right), \ldots, \left(\alpha_{id}, \beta_{id}\right)\right\} = \operatorname{Softplus}(f(\mathbf{A}, x_i|\theta))$, where $(\alpha_{ij}, \beta_{ij})$ is the pair of parameters for the $j$-th Beta distribution, and $\operatorname{Softplus}$ maps each parameter in the Beta embedding to its proper range. Here, we use the graph convolution layer in GCN as our encoder.

Given the node embeddings, we define the disjunction operation to derive class embeddings for known classes. 
Given a set of $n$ elements $\left\{\mathcal{Z}_1, \ldots, \mathcal{Z}_n\right\} = \left\{\left(\boldsymbol{\alpha}_1, \boldsymbol{\beta}_1\right), \ldots, \left(\boldsymbol{\alpha}_n, \boldsymbol{\beta}_n\right)\right\}$, in which each element can be either a class or a node in the form of Beta embedding, 
the disjunction operation $\cup_{i=1}^n \mathcal{Z}_i$ is defined as
\begin{equation}
\begin{aligned}
    &\mathbf{DISJUNCTION}\left(\left\{\left(\boldsymbol{\alpha}_1, \boldsymbol{\beta}_1\right),\left(\boldsymbol{\alpha}_{2}, \boldsymbol{\beta}_{2}\right), \ldots, \left(\boldsymbol{\alpha}_{i}, \boldsymbol{\beta}_{i}\right) \right\}\right)\\ &=\operatorname{Softplus}\left(h_2\left(\left(\frac{\sum_{i=1}^nh_1([\boldsymbol{\alpha}_i || \boldsymbol{\beta}_i])}n\right) \cdot \boldsymbol{w} + \textit{bias} \right) \right),
\end{aligned}
\end{equation}
where $\boldsymbol{w}$ is the trainable weight vector, and $h_1$ and $h_2$ are two projectors, and $bias$ is a trainable bias parameter. The reason for applying two projectors $h_1$ and $h_2$ is that Beta embedding is naturally a set of Beta distributions with bounded support. Simply aggregating these distributions or their parameters does not align with their statistical properties. 
Thus, we use two projectors to first project Beta embeddings into a different embedding space where aggregation can be applied and then project the aggregated embedding back to the embedding space of the Beta embedding to obtain the proposition (in the form of  Beta embedding) after disjunction. 

We then represent each known class as a disjunction of training nodes from this class. Mathematically, given a set of $n$ training nodes $\{\mathcal{N}_{1k}, \ldots, \mathcal{N}_{nk}\}$ from class $k$ with $\mathcal{N}_{ik}$ being the Beta embedding of an arbitrary node $i$, the class embedding for class $k$ is defined as 
\begin{equation}
    \mathcal{C}_k = \mathbf{DISJUNCTION}\left(\left\{\mathcal{N}_{1k}, \ldots, \mathcal{N}_{nk}\right\}\right).
\end{equation}
Figure~\ref{fig:modules}(a) presents an illustrative example of the class embeddings reasoned from the disjunction operation, i.e., the blue and green shaded areas represent class embeddings of two known classes (blue and green). Then, we integrate context information to avoid overconfidence by using a class-specific graph convolutional network (GCN) to compute the evidence that supports node $i$ belonging to class $k$. Mathematically, for any known class $k \in {1, \ldots, K}$ from a set of $K$ known classes, the evidence of node $i$ belonging to class $k$ is computed as
\begin{equation}
    e_{ik} = \left[\textit{GCN}_k\left(\mathbf{A}, \mathbf{X}_k'\right)\right]_i,
    \label{eq: evidence calculation}
\end{equation}
where $[\cdot]_i$ denotes the $i$-th element of the output vector, $\textit{GCN}_k$ is a class-specific GCN for class $k$, and $\mathbf{X}_k$ is a specially constructed feature matrix that concatenates the node embeddings $\{\mathcal{N}_{1k}, \ldots, \mathcal{N}_{nk}\}$ with the class embeddings $\mathcal{C}_k$ of class $k$. The context information integration allows each node to generate a context-aware embedding for every class and uses it to compute the dissonance score of the node belonging to a specific class. Specifically, we compute the dissonance score for node $i$ as

\begin{equation}
    \begin{aligned}
    \operatorname{Dissonance}_i&=\sum_{k=1}^K\left(\frac{b_{ik} \sum_{j \neq k} b_{ij} \operatorname{Bal}\left(b_{ij}, b_{ik}\right)}{\sum_{j \neq k} b_{ij}}\right),\\
    \end{aligned}
    \label{eq: uncertainty calculation}
\end{equation}
where $b_{ik} = e_{ik}/S_i$, $S_i=\sum_{k=1}^K e_{ik} + W_i$, and $\operatorname{Bal}\left(b_k, b_j\right) = 1-\left|b_k-b_j\right| /\left(b_k+b_j\right)$ is the relative mass balance between a pair of masses $b_k$ and $b_j$. A key difference between the subjective logic in \name~and the traditional subjective logic approaches is that $W_i$ is a trainable prior weight that accounts for potentially novel classes (Eq.~\eqref{eq:trainable-prior-weight}). We defer detailed computation of $W_i$ to Section~\ref{subsec:vacuity-reasoning}. 

\subsection{M2. Characterizing Novel Classes via Vacuity Reasoning}\label{subsec:vacuity-reasoning}
Traditional OOD detection techniques typically do not model the representations of novel OOD class(es) explicitly. It might be suboptimal when the boundary between in-distribution data and out-of-distribution data is not clearly defined in high-dimensional space. In contrast, our proposed \name~introduces an explicit support region~\cite{Awad2015} for known classes and an implicit support region for novel classes through logical reasoning, which could yield better capability for OOD detection. 
Specifically, to represent the support regions for known and novel classes, we define the support region of all known classes as the disjunction of class embeddings obtained via Dissonance Reasoning
\begin{equation}
    \mathcal{C}_\text{Known}=\cup_{k=1}^K \mathcal{C}_k = \mathbf{DISJUNCTION}(\{\mathcal{C}_1, \ldots, \mathcal{C}_K\}).
\end{equation}

Then the support region of novel class(es) $\mathcal{C}_\text{Nov}$ is defined as the logical negation of the support region of all known classes
\begin{equation}
    \mathcal{C}_\text{Nov} = \overline{\cup_{k=1}^K \mathcal{C}_k},
\end{equation}
where the negation operation is defined as 
\begin{equation}
    \mathbf{NEGATION}(\boldsymbol{\alpha}_i, \boldsymbol{\beta}_i)=(\frac{1}{\boldsymbol{\alpha}_i},\frac{1}{\boldsymbol{\beta}_i}).
\end{equation}
The intuition is that if a node belongs to a novel class, it certainly should not belong to any of the known classes. As shown in Figure~\ref{fig:modules}(b), this approach offers an implicit support region of the novel class(es).

Similar to the context information integration in (M1) Dissonance Reasoning, we compute a context-aware prior weight $W_i$ for any node $i$ in the graph 
\begin{equation}\label{eq:trainable-prior-weight}
    W_i = [\textit{GCN}_\text{Nov}(\mathbf{A}, \mathbf{X}_\text{Nov})]_i,
\end{equation}
where $[\cdot]_i$ denotes the $i$-th element of the output vector, and $\mathbf{X}_\text{Nov}$ is a specially constructed feature matrix that concatenates the node embeddings with the class embedding of the novel class $\mathcal{C}_\text{Nov}$. Then we compute the vacuity score for node $i$, which indicates how likely the node $i$ is to belong to novel class(es), as
\begin{equation}
    \operatorname{Vacuity}_i=\frac{W_i}{S_i},
\end{equation}
where $S_i=\sum_{k=1}^K e_{ik} + W_i$, and $e_{ik}$ is the evidence that supports node $i$ belonging to class $k$ (Eq.\eqref{eq: evidence calculation}). 

\noindent \textbf{Remark:} Traditional subjective logic frameworks preset the prior weight $W$ to the number of classes, which could lead to overconfidence~\cite{chen2024redl}. For example, high confidence could be assigned to misclassified samples, and the model could make high-confidence predictions that classify out-of-distribution data to in-distribution known classes~\cite{chen2024redl}. The main problem is that the fixed prior weight disallows the model to adjust its uncertainty based on the observed evidence, which causes inaccurate uncertainty estimates. \name~overcomes this limitation by treating the prior weight $W_i$ for each node as a learnable parameter for better capabilities in both misclassification detection tasks and OOD detection.

Besides the dissonance score and vacuity score, we also
compute the probability that node $i$ belongs to class $k$ for in-distribution classification as 
\begin{equation}
    p_{ik} = b_{ik} + a_k\frac{W_i}{S_i},
\label{eq:projected probability}
\end{equation}
where $a_k = 1/K$ is the base rate for any class $k$ in the set of $K$ known classes following the common practice.

\subsection{Optimization}\label{sec:Optimization}

Our training aims to ensure the effective learning of robust Beta embeddings for nodes, the disjunction operation for logical reasoning, and the estimation of class-specific evidence based on this reasoning, ultimately providing a reliable estimation of uncertainty. The whole training is separated into two phases. In the first phase, we focus on ensuring training the encoder for encoding node into Beta embedding and disjunction operation using the idea that the Beta embedding of a node aligns closely with its corresponding class embedding while being distinct from embeddings of other classes. We achieve this by minimizing the KL Divergence between a node's Beta embedding $\mathcal{N}$ and its corresponding class-level Beta embedding $\mathcal{C}$ and maximizing the KL Divergence between other class-level Beta embeddings. Since the encoder is responsible for generating node embeddings and the disjunction operation derives class embeddings from these node embeddings, this method can effectively train both the encoder and the disjunction operation. This loss function $BL$ in this phase for node $i$ with label $y_i$ can be formalized as:
\begin{equation}
    \begin{aligned}
    BL_i=&-\log \sigma(\gamma-\operatorname{Dist}(\mathcal{N}_i ; \mathcal{C}_{y_i})) \\
    &
    -\sum_{\substack{k=1 \\ k \neq y_i}}^{K} \frac{1}{K} \log \sigma\left(\operatorname{Dist}\left(\mathcal{N}_i ; \mathcal{C}_k\right)-\gamma\right), \\
    \end{aligned}
    \label{eq: Beta Loss}
\end{equation}
where $\mathcal{N}_i$ is the Beta embedding for node $i$ belongs to the class $y_i$ with class-level Beta embedding $\mathcal{C}_{y_i}$, $\gamma$ denotes the margin, and Dist is defined as
\begin{equation}
\operatorname{Dist}(\mathcal{N} ; \mathcal{C})=\sum_{j=1}^{d} \mathrm{KL}\left(\left(\alpha_{\mathcal{N}j}, \beta_{\mathcal{N}j}\right) ; \left(\alpha_{\mathcal{C}j}, \beta_{\mathcal{C}j}\right)\right).
\end{equation}
Here $\left(\alpha_{\mathcal{N}j}, \beta_{\mathcal{N}j}\right)$ or $\left(\alpha_{\mathcal{C}j}, \beta_{\mathcal{C}j}\right)$ represents the j-th beta distribution and $d$ is the dimension of a Beta embedding. We use $\operatorname{Dist}(\mathcal{N}; \mathcal{C})$ rather than $\operatorname{Dist}(\mathcal{C}; \mathcal{N})$ to ensure that class embeddings effectively "cover" the modes of all node embeddings belonging to that class. By including the embeddings of other classes (including novel classes) as negative samples, we optimize the model to distinguish the target class from others.

The second phase of the training focuses on training the evidence estimator to estimate the evidence supports that a given node belongs to a specific class for both prediction and uncertainty estimation. To achieve that, we incorporate an expected cross-entropy loss based on the Dirichlet distribution $D(p_{ik}|{\xi}_{ik})$ as the loss function where $p_{ik}$ is the predicted probability indicating that sample $i$ belongs to class $k$ described in Eq. \ref{eq:projected probability} and $\xi_{ik}$ is the total strength of the belief of class $k$ for sample $i$. For a given sample $i$, the loss function $DL$ in phase 2 is expressed as
\begin{equation}
    \begin{aligned}
    DL_i& =\int\left[\sum_{k=1}^K-y_{i k} \log \left(p_{i k}\right)\right] \frac{1}{B\left({\xi}_i\right)} \prod_{k=1}^K p_{i k}^{\xi_{i k}-1} d \mathbf{p}_i  \\
    &=\sum_{k=1}^K y_{i k}\left(\psi\left(S_i\right)-\psi\left(\xi_{i k}\right)\right), \\
    \end{aligned}
\end{equation}
where $\psi$ is the digamma function, $S_i=\sum_{k=1}^K e_{ik} + W_i$ and $\xi_{ik}=e_{ik} + a_kW_i$. $a_k$ is the base rate for class $k$ defined in subjective logic and set as $\frac{1}{K}$ for every class following the common practice.

Unlike traditional subjective-logic-based frameworks that often include a variance-minimization regularization term, which can lead to overconfidence in both misclassification and OOD detection (as shown in \cite{chen2024redl}), we avoid this in our approach.

To enhance the model’s performance further, we adopt an alternating training strategy. This approach iteratively refines both node Beta embeddings and the disjunction operation while improving the evidence estimator and uncertainty estimation. By alternating between these two phases, the model gains a more comprehensive understanding of the data, leading to improved performance.

\section{Experiments}

In this section, we evaluate the performance of \name~on five benchmark datasets to understand the performance of \name~in in-distribution classification, misclassification detection, and out-of-distribution detection. \name~exhibits superior performance compared to various state-of-the-art baselines (Section \ref{sec:overall performance}). Moreover, we conduct ablation studies (Section \ref{sec:ablation study}) to demonstrate the necessity of each module in \name~and report the parameter sensitivity, scalability and complexity in Section \ref{sec:parameter sensitivity}, Appendix \ref{sec:scalability analysis} and Appendix \ref{sec:complexity_comparison} to show that \name~ is scalable and achieves a convincing performance with minimal tuning efforts.

\subsection{Experimental Settings}

\noindent{\textbf{Datasets.}} We consider node classification and evaluate \name~on five benchmark datasets, including Wiki-CS~\cite{mernyei2022wikicswikipediabasedbenchmarkgraph}, Amazon-Photo, Amazon-Computer, Coauthor-CS, and Coauthor-Physics~\cite{shchur2019pitfallsgraphneuralnetwork}. For in-distribution data, we follow the common practice \cite{kipf2017semisupervised} and use random splits with 1:1:8 for training/validation/testing. To create an OOD set for out-of-distribution detection, we follow a label leave-out strategy by considering nodes with certain classes as in-distribution and nodes with other labels as out-of-distribution~\cite{liu2021energybasedoutofdistributiondetection}. Detailed statistics of datasets are summarized 
in Appendix~\ref{sec:dataset stat}.

\noindent{\textbf{Baseline Methods.}} We compare \name~with nine baseline methods in three categories, which include methods that are designed for OOD detection specifically, to misclassification specifically, and to both OOD detection and misclassification detection.

\begin{itemize}[leftmargin=*]

\item \textbf{Methods for OOD detection} include (1) MaxLogit~\cite{hendrycks2022scalingoutofdistributiondetectionrealworld}, which uses the maximum logit as the OOD score, 
(2) ODIN~\cite{liang2020enhancingreliabilityoutofdistributionimage}, which improves OOD detection with small perturbations to the input data and temperature scaling, 
(3) Mahalanobis~\cite{lee2018simpleunifiedframeworkdetecting}, which uses a confidence score based on the Mahalanobis distance for OOD detection, (4) Energy~\cite{liu2021energybasedoutofdistributiondetection} which detects OOD data by an energy-based discriminator and (5) GNNSafe~\cite{wu2023energybasedoutofdistributiondetectiongraph}, which is a graph-specific method and also detects OOD data by an energy-based discriminator.
To evaluate the performance of misclassification detection of these methods, we use their OOD score as the misclassification score for misclassification detection. 

\item \textbf{Methods for misclassification detection} include 
(1) CRL~\cite{moon2020confidenceawarelearningdeepneural}, which improves confidence estimates by optimizing the correctness ranking loss,
and (2) Doctor~\cite{NEURIPS2021_2cb6b103}, which uses the probabilistic output by the model to determine whether we should trust model prediction or not. 
To evaluate the performance of OOD detection of these methods, we use their misclassification score as the OOD score for OOD detection.

\item \textbf{Methods for both OOD detection and misclassification detection} include 
(1) GPN~\cite{stadler2021graphposteriornetworkbayesian}, which estimates the uncertainty of model predictions with Bayesian posterior updates, 
and (2) GKDE~\cite{NEURIPS2020_968c9b4f}, which estimates node uncertainty by leveraging a Dirichlet prior based on graph-based kernel distribution estimation. 
Following existing settings~\cite{stadler2021graphposteriornetworkbayesian, NEURIPS2020_968c9b4f}, we use aleatoric uncertainty estimate for misclassification detection and epistemic uncertainty estimate for OOD detection.
\end{itemize}

\noindent{\textbf{Evaluation Metrics.}} We evaluate the performance of \name~for in-distribution classification, misclassification detection, and OOD detection. For in-distribution classification, we calculate the classification accuracy (Acc) of in-distribution test samples. For misclassification detection, we calculate the area under the risk coverage curve (AURC), which measures the error rate using samples whose confidence is higher than a pre-defined confidence threshold \cite{geifman2017selectiveclassificationdeepneural}. For OOD detection, we calculate the false positive rate at 95\% true positive rate (FPR95), which quantifies the probability that an OOD example is predicted as an in-distribution when the true positive rate is 95\%, and the area under the receiver operating characteristic curve (AUROC). We perform 5 runs and report the mean and standard deviation of each metric in our experiments.

\noindent{\textbf{Implementation Details.}} We set the encoder in \name~to be two GCN layers, each of which has 64 hidden channels and is followed by a batch normalization layer and a softplus function. For a fair comparison, we apply the same encoder to MaxLogit, ODIN, CRL, Doctor, Mahalanobis, Energy, and GNNSafe. The main reasons are twofold: (1) MaxLogit, ODIN, CRL, Doctor, Mahalanobis, and Energy are not specifically designed for graph data, and (2) GNNSafe is a generic framework that can be applied to any encoder. For GPN and GKDE, we follow their publicly released implementations and use hyperparameters reported in the original paper. Regarding model selection, we select the model with the best overall performance on the validation set, where the overall performance is calculated by $\text{Acc} + \text{AUROC} - 10*\text{AURC}$. More details on hyperparameter settings for the learning rate, dropout rate, and $\gamma$ are deferred to Appendix~\ref{appendix:hyperparam-setting}.

\subsection{Effectiveness Results}\label{sec:overall performance}

\begin{table*}[t]
\setlength{\tabcolsep}{1pt}
\centering
\caption{Effectiveness of in-distribution classification (IDC), misclassification detection (MD), and out-of-distribution detection (OODD) on Amazon-photo, Amazon-computer, Coauthor-cs, Coauthor-physics, and Wiki-cs. We report the mean and standard deviation for each method on each dataset over five runs. AURC is multiplied by 1000, all other metrics are in percentage. Lower is better for $\downarrow$, and higher is better for $\uparrow$. Bold and \textcolor{red}{red} indicate best, and underlined and \textcolor[rgb]{0,0.5,0}{green} indicate second best.}
\resizebox{1\textwidth}{!}{
\renewcommand{\arraystretch}{1.2}
\begin{tabular}{c|c|c|ccccc|cc|cc|c}
\hline
Dataset & \multicolumn{2}{c|}{Metric} & ODIN & MaxLogit & Mahalanobis & Energy & GNNSafe & CRL & DOCTOR & GKDE & GPN & \name \\ \hline
\hline
\multirow{4}{*}{\makecell{Amazon\\-photo}} & IDC & Acc ($\uparrow$) & $92.59 \pm 1.92$ & $94.75 \pm 0.32$ & $90.07 \pm 4.15$ & $94.06 \pm 0.91$ & $94.35 \pm 0.43$ & $93.18 \pm 0.78$ &  \textcolor{red}{\boldmath$\mathbf{94.90 \pm 0.36}$} & $94.27 \pm 1.18$ & $91.21 \pm 2.98$ & \textcolor[rgb]{0,0.5,0}{\underline{$94.76 \pm 0.24$}} \\ \cline{2-3}
& MD & AURC ($\downarrow$) &  $25.48 \pm 6.79$ & $14.03 \pm 1.20$ & $129.50 \pm 46.33$ & $16.56 \pm 2.13$ & \textcolor[rgb]{0,0.5,0}{\underline{$10.78 \pm 0.85$}} & $14.70 \pm 2.53$ & $13.64 \pm 2.35$ & $12.76 \pm 3.41$ & $20.97 \pm 10.74$ &  \textcolor{red}{\boldmath$\mathbf{6.99 \pm 0.36}$} \\  \cline{2-3}
& \multirow{2}{*}{OODD} & FPR95 ($\downarrow$) & $42.84 \pm 23.18$ & $7.69 \pm 2.36$ & $69.87 \pm 8.42$ & $6.53 \pm 2.20$ &  \textcolor{red}{\boldmath$\mathbf{4.43 \pm 2.91}$} & $15.81 \pm 6.03$ & $13.81 \pm 4.78$ & $37.45 \pm 25.45$ & $53.99 \pm 31.03$ & \textcolor[rgb]{0,0.5,0}{\underline{$4.79 \pm 2.08$}} \\ \cline{3-3}
& & AUROC ($\uparrow$) & $91.44 \pm 4.82$ & $98.04 \pm 0.43$ & $70.45 \pm 4.83$ & \textcolor[rgb]{0,0.5,0}{\underline{$98.20 \pm 0.47$}} & $97.88 \pm 0.31$ & $97.07 \pm 0.67$ & $97.22 \pm 0.66$ & $91.74 \pm 4.76$ & $85.96 \pm 8.31$ &  \textcolor{red}{\boldmath$\mathbf{98.50 \pm 0.34}$} \\ \cline{2-3}
\hline
\hline
\multirow{4}{*}{\makecell{Amazon\\-computer}} & IDC & Acc ($\uparrow$) & $74.18 \pm 19.20$ & $86.79 \pm 1.74$ & $76.62 \pm 10.61$ & $86.05 \pm 2.11$ & \textcolor[rgb]{0,0.5,0}{\underline{$87.55 \pm 1.38$}} & $84.73 \pm 0.72$ & $86.84 \pm 1.15$ & $83.58 \pm 4.75$ & $78.20 \pm 14.49$ &  \textcolor{red}{\boldmath$\mathbf{87.73 \pm 0.20}$} \\ \cline{2-3}
 & MD & AURC ($\downarrow$) & $183.05 \pm 202.77$ & $61.46 \pm 5.07$ & $272.28 \pm 48.16$ & $66.51 \pm 7.99$ & $58.31 \pm 4.95$ & \textcolor[rgb]{0,0.5,0}{\underline{$50.65 \pm 2.03$}} & $52.04 \pm 6.71$ & $73.25 \pm 42.03$ & $80.97 \pm 42.42$ &  \textcolor{red}{\boldmath$\mathbf{31.42 \pm 1.88}$} \\ \cline{2-3}
 & \multirow{2}{*}{OODD} & FPR95 ($\downarrow$) & $54.43 \pm 21.23$ & $37.04 \pm 10.95$ & $84.04 \pm 4.56$ & $35.91 \pm 10.24$ & \textcolor[rgb]{0,0.5,0}{\underline{$34.17 \pm 7.60$}} & $40.84 \pm 4.23$ & $38.98 \pm 7.60$ & $70.01 \pm 16.37$ & $74.99 \pm 19.39$ &  \textcolor{red}{\boldmath$\mathbf{22.01 \pm 3.27}$} \\ \cline{3-3}
 & & AUROC ($\uparrow$) & $81.58 \pm 15.67$ & $90.99 \pm 2.87$ & $65.63 \pm 2.58$ & $91.18 \pm 1.43$ & \textcolor[rgb]{0,0.5,0}{\underline{$91.20 \pm 2.48$}} & $90.24 \pm 1.31$ & $90.17 \pm 2.45$ & $79.61 \pm 7.59$ & $79.94 \pm 11.73$ &  \textcolor{red}{\boldmath$\mathbf{95.01 \pm 0.76}$} \\ \cline{2-3}
\hline
\hline
\multirow{4}{*}{\makecell{Coauthor\\-cs}} & IDC & Acc ($\uparrow$) & $91.87 \pm 2.20$ & $94.31 \pm 0.43$ & $91.07 \pm 3.08$ & $93.69 \pm 1.20$ & \textcolor[rgb]{0,0.5,0}{\underline{$94.31 \pm 0.38$}} & $92.00 \pm 1.52$ & $93.98 \pm 0.69$ &  \textcolor{red}{\boldmath$\mathbf{95.00 \pm 0.15}$} & $85.57 \pm 6.92$ & $93.88 \pm 0.14$ \\ \cline{2-3}
 & MD & AURC ($\downarrow$) & $18.30 \pm 6.29$ & $15.56 \pm 1.80$ & $104.60 \pm 34.84$ & $17.02 \pm 2.23$ & $19.25 \pm 1.11$ & $16.47 \pm 3.50$ & \textcolor[rgb]{0,0.5,0}{\underline{$11.65 \pm 4.20$}} & $16.71 \pm 0.50$ & $35.30 \pm 29.28$ &  \textcolor{red}{\boldmath$\mathbf{8.53 \pm 0.17}$} \\ \cline{2-3}
 & \multirow{2}{*}{OODD} & FPR95 ($\downarrow$) & $31.55 \pm 7.26$ & $19.00 \pm 4.56$ & $88.53 \pm 4.15$ & $17.98 \pm 2.67$ & \textcolor[rgb]{0,0.5,0}{\underline{$12.40 \pm 2.01$}} & $39.38 \pm 12.90$ & $27.83 \pm 8.01$ & $14.76 \pm 0.57$ & $65.15 \pm 20.10$ &  \textcolor{red}{\boldmath$\mathbf{5.82 \pm 0.47}$} \\ \cline{3-3}
 & & AUROC ($\uparrow$) & $93.21 \pm 1.20$ & $95.37 \pm 1.17$ & $59.23 \pm 6.13$ & $95.77 \pm 0.52$ & \textcolor[rgb]{0,0.5,0}{\underline{$96.96 \pm 0.33$}} & $91.79 \pm 2.76$ & $94.02 \pm 0.77$ & $96.80 \pm 0.25$ & $84.39 \pm 5.97$ &  \textcolor{red}{\boldmath$\mathbf{98.53 \pm 0.10}$} \\ \cline{2-3}
\hline
\hline
\multirow{4}{*}{\makecell{Coauthor\\-physics}} & IDC & Acc ($\uparrow$) & $96.27 \pm 0.50$ & $97.59 \pm 0.18$ & $94.90 \pm 2.13$ & \textcolor[rgb]{0,0.5,0}{\underline{$97.71 \pm 0.12$}} & $97.69 \pm 0.18$ & $96.25 \pm 0.73$ & $97.56 \pm 0.10$ &  \textcolor{red}{\boldmath$\mathbf{97.94 \pm 0.08}$} & $87.57 \pm 12.92$ & $97.62 \pm 0.03$ \\ \cline{2-3}
 & MD & AURC ($\downarrow$) & $5.80 \pm 1.77$ & $2.97 \pm 0.54$ & $66.98 \pm 25.08$ & $2.54 \pm 0.57$ & $3.86 \pm 0.73$ & $4.06 \pm 0.81$ & $2.48 \pm 0.47$ & \textcolor[rgb]{0,0.5,0}{\underline{$2.11 \pm 0.38$}} & $19.10 \pm 20.30$ &  \textcolor{red}{\boldmath$\mathbf{1.66 \pm 0.13}$} \\ \cline{2-3}
 & \multirow{2}{*}{OODD} & FPR95 ($\downarrow$) & $18.08 \pm 2.17$ & $8.56 \pm 1.36$ & $93.30 \pm 7.33$ & $8.31 \pm 1.23$ & \textcolor[rgb]{0,0.5,0}{\underline{$7.49 \pm 1.06$}} & $18.59 \pm 2.29$ & $11.24 \pm 1.43$ & $8.51 \pm 0.52$ & $33.70 \pm 25.72$ &  \textcolor{red}{\boldmath$\mathbf{3.30 \pm 0.32}$} \\ \cline{3-3}
 & & AUROC ($\uparrow$) & $95.99 \pm 0.43$ & $97.40 \pm 0.35$ & $49.55 \pm 9.83$ & $97.49 \pm 0.44$ & $97.78 \pm 0.28$ & $95.88 \pm 0.45$ & $96.99 \pm 0.47$ & \textcolor[rgb]{0,0.5,0}{\underline{$98.04 \pm 0.13$}} & $93.37 \pm 4.31$ &  \textcolor{red}{\boldmath$\mathbf{99.18 \pm 0.11}$} \\
\hline
\hline

\multirow{4}{*}{Wiki-cs} & IDC & Acc ($\uparrow$) & $68.38 \pm 6.75$ & $82.25 \pm 0.59$ & $76.91 \pm 2.76$ & \textcolor[rgb]{0,0.5,0}{\underline{$82.63 \pm 0.66$}} &  \textcolor{red}{\boldmath$\mathbf{82.72 \pm 1.35}$} & $82.00 \pm 0.35$ & $82.58 \pm 0.27$ & $82.00 \pm 1.30$ & $79.84 \pm 3.55$ & $82.32 \pm 1.01$ \\ \cline{2-3}
 & MD & AURC ($\downarrow$) & $184.52 \pm 105.42$ & $62.96 \pm 4.33$ & $227.81 \pm 28.97$ & $77.84 \pm 7.60$ & $81.00 \pm 11.33$ & $58.88 \pm 3.70$ & $60.85 \pm 3.21$ & \textcolor[rgb]{0,0.5,0}{\underline{$56.93 \pm 5.66$}} & $60.58 \pm 14.72$ &  \textcolor{red}{\boldmath$\mathbf{49.27 \pm 4.33}$} \\ \cline{2-3}
 & \multirow{2}{*}{OODD} & FPR95 ($\downarrow$) & $84.87 \pm 4.73$ & $66.34 \pm 2.04$ & $91.28 \pm 3.44$ & \textcolor[rgb]{0,0.5,0}{\underline{$63.60 \pm 6.37$}} & $70.43 \pm 2.16$ & $70.24 \pm 4.55$ & $66.91 \pm 2.61$ & $71.91 \pm 2.61$ & $85.23 \pm 2.38$ &  \textcolor{red}{\boldmath$\mathbf{55.20 \pm 7.52}$} \\ \cline{3-3}
 & & AUROC ($\uparrow$) & $68.99 \pm 9.63$ & $84.15 \pm 0.57$ & $54.59 \pm 1.43$ & $83.01 \pm 3.39$ & \textcolor[rgb]{0,0.5,0}{\underline{$84.73 \pm 1.66$}} & $83.72 \pm 1.11$ & $83.78 \pm 0.77$ & $81.77 \pm 1.39$ & $68.71 \pm 2.49$ &  \textcolor{red}{\boldmath$\mathbf{86.95 \pm 3.51}$} \\
\hline
\end{tabular}
}
\label{tab:result1}
\end{table*}

Table~\ref{tab:result1} presents the effectiveness of \name~in the tasks of in-distribution classification, misclassification detection, and out-of-distribution.
In general, we have the following observations: \textbf{(1) In-distribution classification.} In terms of in-distribution classification accuracy, \name~demonstrates consistently competitive performance across all datasets. More specifically, \name~attains the highest accuracy at 87.73\%, slightly surpassing GNNSafe (87.55\%) on Amazon-computer; The accuracy of \name~closely approaches the top-performing baseline method GKDE (97.62\% for \name~vs. 97.94\% for GKDE) on Coauthor-physics. 
\textbf{(2) Misclassification detection.} \name~excels in misclassification detection, consistently achieving the lowest AURC scores across all datasets. For instance, on Amazon-photo, 
\name~could achieve a significantly lower AURC (6.99) compared to the second-best method, GNNSafe (10.78).
This consistent trend across datasets underscores 
the ability of \name~for misclassification detection.
\textbf{(3) Out-of-distribution detection.}
\name~consistently delivers the state-of-the-art performance in both FPR95 and AUROC. For instance, on Amazon-computer, \name~significantly outperforms other methods, achieving an FPR95 of 22.01\% (vs. 34.17\% for GNNSafe, the second-best method) and an AUROC of 95.01\% (vs. 91.20\% for GNNSafe, the second-best method). This strong performance highlights the ability of \name~to detect out-of-distribution data effectively.
\textbf{(4) Overall balanced performance.} While \name~might not always be the top performer in specific tasks on every dataset, its ability to strike an effective balance across tasks leads to consistently strong overall results. For instance, even though the accuracy of \name~on Amazon-photo is 0.15\% lower than the accuracy of DOCTOR (the best-performing method)
, it more than compensates for this with a 48.75\% lower AURC and a 1.32\% higher AUROC. These results demonstrate that \name~effectively balances the trade-offs between accuracy, misclassification detection, and OOD detection, leading to strong overall performance across multiple tasks. Appendix~\ref{sec:additional result} presents additional experimental results to further support our findings.

\vspace{-4mm}
\subsection{Ablation Study}\label{sec:ablation study}

\begin{table}[t]
\caption{Ablation study of \name~on Amazon-photo by including ($\checkmark$) or excluding (M1) the dissonance reasoning module, (M2) the vacuity reasoning module, and (CI) context information integration in the tasks of in-distribution classification (IDC), misclassification detection (MD) and out-of-distribution detection (OODD). We report the mean and standard deviation of each variant over five runs. AURC is multiplied by 1000, all other metrics are in percentage. Lower is better for $\downarrow$ and higher is better for $\uparrow$. Bold and  \textcolor{red}{red} indicate best.}
\setlength{\tabcolsep}{3pt}
\centering
\resizebox{.48\textwidth}{!}{
\begin{tabular}{c|ccc|c|c|cc}
\hline
\multirow{2}{*}{Variant} & \multicolumn{3}{c|}{Components} & IDC & MD & \multicolumn{2}{c}{OODD}\\ 
\cline{2-8}
& M1 & M2 & CI & Acc ($\uparrow$) & AURC ($\downarrow$) & FPR95 ($\downarrow$) & AUROC ($\uparrow$) \\
\hline
(a) & & & & $93.07 \pm0.40$ & $11.56 \pm1.40$ & $44.19 \pm12.60$ &  $91.12 \pm2.11$  \\
(b) & $\checkmark$ & & & $94.55 \pm0.53$ & $8.50 \pm1.40$ & $49.96 \pm11.88$ &  $90.90 \pm1.91$  \\
(c) & $\checkmark$ & $\checkmark$ &  & $94.33 \pm0.33$ & $9.10 \pm1.05$ & $19.92 \pm6.65$ &  $95.59 \pm0.86$  \\
(d) & $\checkmark$ & & $\checkmark$ & $94.43 \pm0.25$ & $7.07 \pm0.32$ & $8.03 \pm2.33$ &  $97.35 \pm0.49$  \\
(e) & $\checkmark$ & $\checkmark$ & $\checkmark$ &  \textcolor{red}{\boldmath${\mathbf{94.76 \pm0.24}}$} &  \textcolor{red}{\boldmath${\mathbf{6.99 \pm0.36}}$} &  \textcolor{red}{\boldmath${\mathbf{4.79 \pm2.08}}$} &  \textcolor{red}{\boldmath${\mathbf{98.50 \pm0.34}}$} \\
\hline
\end{tabular}
}
\label{tab:module ablation}
\end{table}

Table \ref{tab:module ablation} presents five variants of \name~and their performance on Amazon-photo, including 
(a) only subjective logic uncertainty estimation with a GCN generating evidence directly, (b) \name\ without Vacuity Reasoning (M2) and context information integration (CI), using MLP for evidence calculation and fixed prior weight W, (c) \name\ without CI, using MLP instead of GCN for evidence calculation, (d) \name\ without M2, using fixed prior weight W for each sample, (e) complete \name. From the results, we have several interesting observations: (1) Significant performance improvements are evident in configuration (b) compared with (a), with a 1.59\% boost in accuracy, a 26.47\% reduction in AURC, and a 0.22\% increase in AUROC. Although FPR95 increases slightly by 13.06\%, the gains in overall performance demonstrate the effectiveness of Beta embedding in the Dissonance Reasoning module (M1) for generating representations that capture misclassified data;
(2) The addition of Vacuity Reasoning (M2) in (c) results in minor decreases in accuracy (0.23\%) and a 7.06\% increase in AURC compared with (b), yet it leads to substantial improvements in OOD detection. FPR95 decreases by 60.13\%, and AUROC improves by 5.16\%, indicating that Vacuity Reasoning enhances 
the performance of OOD detection, despite slight trade-offs in in-distribution classification; (3) The inclusion of context information integration (CI) in (d) yields notable benefits, with accuracy improving by 0.12\%, AURC reducing by 16.82\%, FPR95 dropping by 83.93\%, and AUROC increasing by 7.10\% compared with (b). These metrics show that integrating context information significantly enhances the performance in both misclassification and OOD detection; (4) The complete \name\ model demonstrates the synergistic effects of integrating all modules. Compared to the previous best configuration (d), the full model increases accuracy by 0.35\%, reduces AURC by 1.13\%, lowers FPR95 by 40.35\%, and increases AUROC by 1.18\%. These results affirm that combining all components leads to optimal performance across in-distribution classification, misclassification detection, and OOD detection tasks.

\begin{table}[t]
\caption{Ablation study of \name~with (w/) or without (w/o) alternate training (AT) on Amazon-photo in the tasks of in-distribution classification (IDC), misclassification detection (MD), and out-of-distribution detection (OODD). We report the mean and standard deviation of each variant over five runs. AURC is multiplied by 1000, all other metrics are in percentage. Lower is better for $\downarrow$, and higher is better for $\uparrow$. Bold and  \textcolor{red}{red} indicate best.}
\setlength{\tabcolsep}{3pt}
\centering
\begin{tabular}{cc|c|c|cc}
\hline
& \multirow{2}{*}{Method} & IDC & MD & \multicolumn{2}{c}{OODD}\\ \cline{3-6}
 &  & Acc ($\uparrow$) & AURC ($\downarrow$) & FPR95 ($\downarrow$) & AUROC ($\uparrow$) \\
\hline
 & w/o AT & $94.50 \pm0.10$ &  \textcolor{red}{\boldmath${\mathbf{6.90 \pm0.33}}$} & $5.64 \pm2.26$ & $98.29 \pm0.33$ \\\hline
 & w/ AT &  \textcolor{red}{\boldmath${\mathbf{94.76 \pm0.24}}$} & $6.99 \pm0.36$ &  \textcolor{red}{\boldmath${\mathbf{4.79 \pm2.08}}$} &  \textcolor{red}{\boldmath${\mathbf{98.50 \pm0.34}}$} \\
\hline
\end{tabular}
\label{tab:training ablation}
\end{table}

Moreover, we experiment with evaluating the impact of using the alternate training strategy on the Amazon-photo dataset, as shown in Table \ref{tab:training ablation}. With alternate training, we observe improvements in accuracy for in-distribution classification, and notably better performance in out-of-distribution detection as indicated by the lower FPR95 and higher AUROC values. While there is a slight increase in AURC for misclassification detection, the overall benefits in other metrics, especially for OOD detection, suggest that the alternate training strategy contributes to a more robust and balanced model performance when considering both misclassification detection and OOD detection.

\subsection{Parameter Sensitivity Analysis}\label{sec:parameter sensitivity}

\begin{figure}[t]
\centering
\includegraphics[width=1\columnwidth]{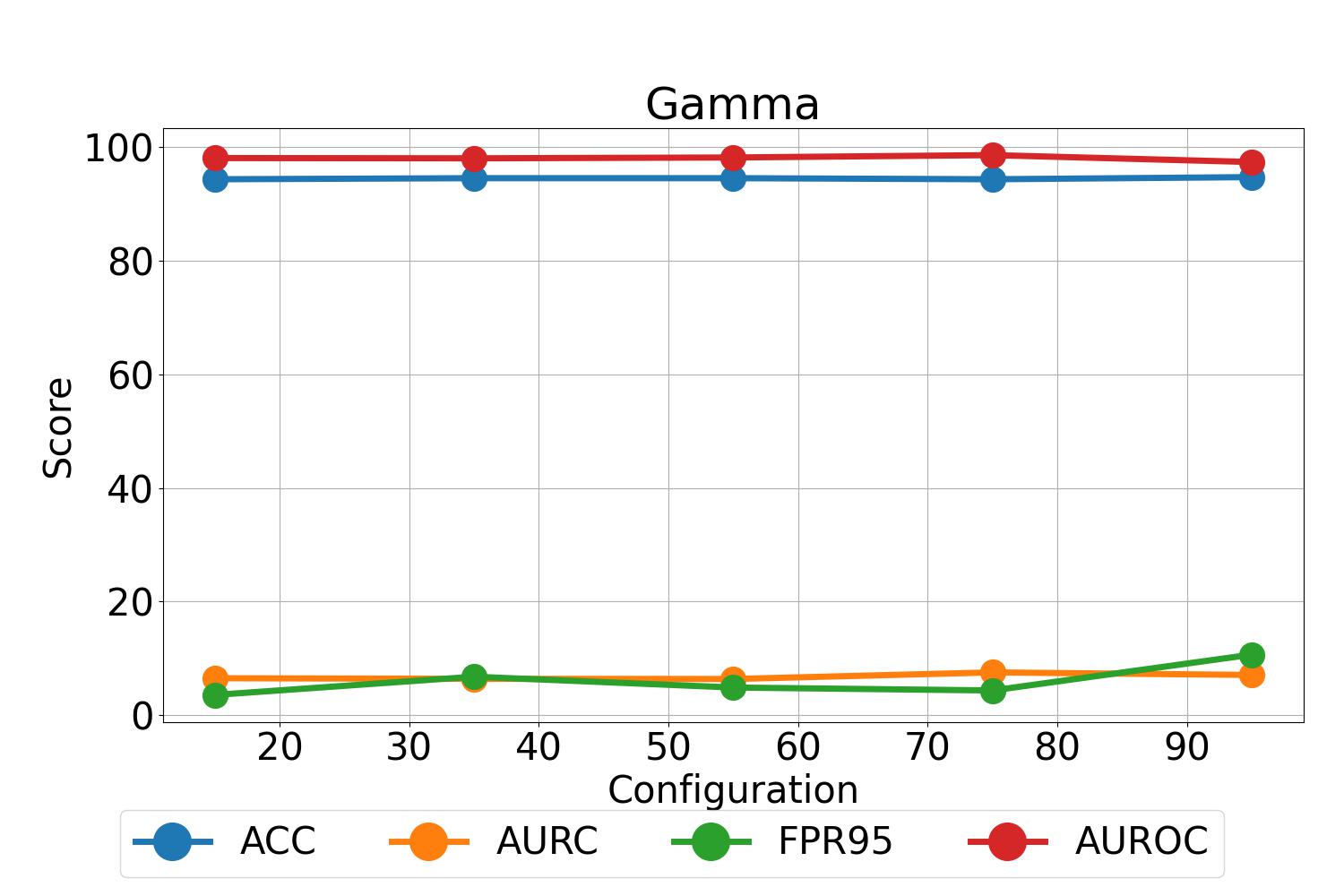}
\caption{Hyperparameter analysis for $\gamma$ on Amazon-photo}
\Description{figure description}
\label{fig:gamma}
\end{figure}

Since the only novel hyperparameter introduced in our model is the margin value $\gamma$  mentioned in Eq. \eqref{eq: Beta Loss}, we study its sensitivity in this section. The analysis results are presented in Figure~\ref{fig:gamma}. From the figure, we observe that \name~exhibits high stability in terms of accuracy (Acc), AURC, FPR95, and AUROC across different values of $\gamma$, demonstrating that \name~is generally robust to $\gamma$.

\noindent \textbf{Remark:} We also present the analysis of the relationship between the running time and the number of nodes and edges, and the computational overhead and memory usage in Appendix~\ref{sec:scalability analysis} due to space limitation.

\section{Related Work}

\textbf{Out-of-distribution (OOD) detection} aims 
to distinguish between in-distribution samples from known classes and OOD samples from unknown/novel classes, which is critical to several real-world applications. In cybersecurity, it plays a key role in detecting unfamiliar cyber threats \cite{10129848, 10154410, electronics12020323}, while in healthcare, it helps distinguish between common and rare diseases, improving diagnostic accuracy \cite{9557572, CAO2023102451}. OOD detection also has notable applications in signal processing, aiding in differentiating known and unknown signal patterns \cite{openfacesurvey, 9163102, Gunther_2017_CVPR_Workshops, Zhao_2023_CVPR}. Representative approaches include ODIN \cite{liang2020enhancingreliabilityoutofdistributionimage}, which enhances softmax scores, Mahalanobis-based methods \cite{lee2018simpleunifiedframeworkdetecting}, utilizing distance metrics, and Energy-based OOD detection \cite{liu2021energybasedoutofdistributiondetection}, leveraging energy functions in neural networks. These methods show strong performance in 
preventing high-confidence but wrong predictions for 
OOD samples. 
Regarding OOD detection on graphs, techniques like GPN \cite{stadler2021graphposteriornetworkbayesian} apply Graph Neural Networks (GNNs) to estimate uncertainty to detect OOD data. GNNSafe~\cite{wu2023energybasedoutofdistributiondetectiongraph} detects OOD nodes by using an OOD discriminator based on an energy function defined with the logits of graph neural networks.

\noindent\textbf{Misclassification detection}, in contrast to OOD detection, primarily addresses the problem of predicting in-distribution samples to a wrong known classes, focusing especially on instances at the boundaries of class categories or those involving noise.
\cite{vazhentsev-etal-2022-uncertainty} presents two computationally efficient methods to estimate uncertainty in Transformer-based models, addressing misclassification detection in named entity recognition and text classification tasks, which are comparable or even better performance to more computationally intensive methods.
DOCTOR \cite{NEURIPS2021_2cb6b103} verifies the predictions made by deep neural networks with consideration of varying level of access to the model to achieve state-of-the-art performance in misclassification without requiring knowledge of the dataset. 

Though both misclassification detection and OOD detection are actively studied, there is limited attention to addressing both tasks on graphs. To our best knowledge, only GPN~\cite{stadler2021graphposteriornetworkbayesian} and GKDE~\cite{NEURIPS2020_968c9b4f} evaluate the performance in both tasks. Our work differs from GPN in that we quantify node uncertainty using Beta embedding with subjective logic, instead of Bayesian uncertainty estimation. Compared to GKDE, we avoid the overconfidence caused by the fixed prior weight and variance-minimizing regularization by introducing a learnable prior weight and the alternate training strategy.

\section{Conclusion}
In this paper, we introduce the Evidential Reasoning Network (\name) 
to address misclassification and out-of-distribution detection on graphs. \name~learns node embeddings and class embeddings using Beta embedding to achieve better reasoning ability and to model the support regions of known classes and novel classes. It further applies the subjective logic framework to detect misclassified data by the Dissonance Reasoning module and out-of-distribution data by the Vacuity Reasoning module. Extensive experiments on five widely used benchmark datasets demonstrate that \name\ outperforms state-of-the-art baseline methods in in-distribution classification, misclassification detection, and out-of-distribution detection. This paper highlights the necessity of reasoning and uncertainty quantification on graphs and presents a step forward in graph learning in open and noisy environments.

\section{Acknowledgements}
We thank the anonymous reviewers for their constructive comments. This work is supported by the National Science Foundation under Award No. IIS-2339989 and No. 2406439, DARPA under contract No. HR00112490370 and No. HR001124S0013, U.S. Department of Homeland Security under Grant Award No. 17STCIN00001-08-00,  Amazon-Virginia Tech Initiative for Efficient and Robust Machine Learning, Amazon AWS, Google, Cisco, 4-VA, Commonwealth Cyber Initiative, National Surface Transportation Safety Center for Excellence, and Virginia Tech. The views and conclusions are those of the authors and should not be interpreted as representing the official policies of the funding agencies or the government.

\bibliographystyle{ACM-Reference-Format}
\bibliography{sample-base}

\appendix

\section{Symbols and Notations}\label{sec:notation}

Table~\ref{tab:symbols} presents key symbols and their corresponding notations used in the paper.

\begin{table}[h]
    \centering
    \caption{Key symbols and their descriptions.}
    \setlength{\tabcolsep}{1pt}
    \begin{tabular}{cc}
        \toprule
        \textbf{Symbol} & \textbf{Description} \\
        \midrule
        $\mathcal{V}$ & A node set for a graph. \\
        $\mathcal{E}$ & An edge set for a graph. \\ 
        $\mathbf{A}$ & The adjacency matrix of a graph. \\
        $\mathbf{X}$ & The node feature matrix of a graph. \\
        $\mathcal{Y}$ & The label set of known classes of a graph. \\
        $K$ & The number of known classes. \\ 
        $\mathcal{N}$ & The node Beta embeddings. \\
        $\mathcal{C}$ & The class Beta embeddings. \\
        $\gamma$ & The margin for training Beta embedding. \\ 
        $d$ & The number of Beta distributions. \\ 
        $(\alpha, \beta)$ & a pair of parameters for a Beta distribution. \\
        $\xi$ & the parameter of Dirichlet distribution. \\ 
        $b_{ik}$ & the belief indicates sample $i$ belongs to class k. \\ 
        $e_{ik}$ & the evidence supports sample $i$ belongs to class k. \\ 
        $W_i$ & the prior weight for sample $i$ defined by subjective logic \\ 
        \bottomrule
    \end{tabular}
    \label{tab:symbols}
\end{table}

\begin{table}[h]
\centering
\caption{Dataset statistics. \# Classes (5th column) indicates the number of all classes in the dataset that includes both in-distribution classes and out-of-distribution classes, and \# OOD Classes (6th column) indicates the number of OOD classes only.}
\resizebox{\columnwidth}{!}{
\begin{tabular}{cccccc}
  \hline
  Dataset & \# Nodes  & \# Edges & \# Attributes & \# Classes & \# OOD Classes \\
  \hline
  Amazon-photo  & 7,650  & 238,162 & 745 & 8 & 4  \\
  Amazon-computer & 13,752 & 491,722 & 767 & 10 & 3\\
  Coauthor-cs & 18,333 & 163,788 & 6,805 & 15 & 4 \\
  Coauthor-physics & 34,493 & 495,924 & 8,415 & 5 & 2 \\
  Wiki-cs & 11,701 & 431,726 & 300 & 10 & 3 \\
  \hline
\end{tabular}
}
\label{table:Dataset}
\end{table}

\begin{table*}[t]
\setlength{\tabcolsep}{1pt}
\centering
\caption{Effectiveness of in-distribution classification (IDC), misclassification detection (MD), and out-of-distribution detection (OODD) on Amazon-photo using models selected by validation accuracy. We report the mean and standard deviation for each method on each dataset over five runs. AURC is multiplied by 1000, all other metrics are in percentage. Lower is better for $\downarrow$, and higher is better for $\uparrow$. Bold and \textcolor{red}{red} indicate best, and underlined and \textcolor[rgb]{0,0.5,0}{green} indicate second best.}
\resizebox{1\textwidth}{!}{
\renewcommand{\arraystretch}{1.2}
\begin{tabular}{c|c|c|ccccc|cc|cc|c}
\hline
Dataset & \multicolumn{2}{c|}{Metric} & ODIN & MaxLogit & Mahalanobis & Energy & GNNSafe & CRL & DOCTOR & GKDE & GPN & \name \\ \hline
\hline
\multirow{7}{*}{\makecell{Amazon\\-photo}} 
& IDC & Acc ($\uparrow$) 
& $77.29 \pm 16.53$ & \textcolor[rgb]{0,0.5,0}{\underline{$95.18 \pm 0.24$}} & $86.28 \pm 13.52$ & $95.06 \pm 0.40$ & \textcolor{red}{\boldmath$\mathbf{95.32 \pm 0.23}$} & $94.26 \pm 0.61$ & $95.20 \pm 0.12$ & $94.23 \pm 1.66$ & $90.98 \pm 3.78$ & $94.74 \pm 0.23$ \\ \cline{2-3}

& \multirow{3}{*}{MD} & AURC ($\downarrow$) 
& $101.95 \pm 113.91$ & $20.16 \pm 4.09$ & $167.11 \pm 133.28$ & $18.33 \pm 3.12$ & $15.12 \pm 7.05$ & \textcolor[rgb]{0,0.5,0}{\underline{$13.41 \pm 2.60$}} & $14.61 \pm 3.71$ & $15.88 \pm 2.83$ & $23.81 \pm 14.87$ & \textcolor{red}{\boldmath$\mathbf{7.77 \pm 0.39}$} \\ \cline{3-3}

& & AUROC ($\uparrow$) 
& $81.19 \pm 3.66$ & $77.30 \pm 3.19$ & $41.60 \pm 5.50$ & $79.33 \pm 3.26$ & $80.33 \pm 6.67$ & \textcolor[rgb]{0,0.5,0}{\underline{$87.04 \pm 2.15$}} & $83.92 \pm 2.55$ & \textcolor{red}{\boldmath$\mathbf{87.33 \pm 2.75}$} & $85.53 \pm 2.10$ & $78.96 \pm 1.88$ \\ \cline{3-3}

& & AUPR ($\uparrow$) 
& $91.93 \pm 8.31$ & $98.03 \pm 0.41$ & $83.48 \pm 13.30$ & $98.22 \pm 0.32$ & $98.55 \pm 0.73$ & \textcolor{red}{\boldmath$\mathbf{98.77 \pm 0.29}$} & \textcolor[rgb]{0,0.5,0}{\underline{$98.60 \pm 0.39$}} & $98.63 \pm 0.18$ & $97.95 \pm 1.23$ & $98.22 \pm 0.16$ \\ \cline{2-3}

& \multirow{3}{*}{OODD} & FPR95 ($\downarrow$) 
& $65.12 \pm 19.18$ & $22.29 \pm 21.06$ & $77.15 \pm 9.08$ & $12.09 \pm 4.77$ & \textcolor[rgb]{0,0.5,0}{\underline{$5.97 \pm 4.22$}} & $37.28 \pm 17.72$ & $15.33 \pm 1.53$ & $52.02 \pm 33.53$ & $55.22 \pm 26.59$ & \textcolor{red}{\boldmath$\mathbf{4.52 \pm 1.25}$} \\ \cline{3-3}

& & AUROC ($\uparrow$) 
& $81.38 \pm 20.14$ & $94.96 \pm 4.64$ & $66.92 \pm 3.42$ & $97.03 \pm 0.87$ & \textcolor[rgb]{0,0.5,0}{\underline{$97.58 \pm 0.59$}} & $93.88 \pm 2.65$ & $96.80 \pm 0.33$ & $86.97 \pm 8.70$ & $85.69 \pm 6.72$ & \textcolor{red}{\boldmath$\mathbf{98.39 \pm 0.30}$} \\ \cline{3-3}

& & AUPR ($\uparrow$) 
& $82.13 \pm 19.99$ & $94.22 \pm 4.34$ & $58.84 \pm 2.49$ & $96.05 \pm 1.09$ & \textcolor[rgb]{0,0.5,0}{\underline{$97.81 \pm 0.52$}} & $93.80 \pm 2.61$ & $95.93 \pm 0.54$ & $85.69 \pm 6.75$ & $85.72 \pm 4.18$ & \textcolor{red}{\boldmath$\mathbf{98.25 \pm 0.25}$} \\ \cline{2-3}
\hline
\end{tabular}
}
\label{tab:result2}
\end{table*}

\section{Dataset Statistics}\label{sec:dataset stat}
Table~\ref{table:Dataset} summarizes statistics for the datasets, including the number of nodes (\# Nodes), the number of edges (\# Edges), the number of attributes for each node (\# Attributes), the number of all classes (\# Classes, including known classes and OOD classes), and the number of OOD classes considered in our experiments (\# OOD Classes).

\section{Hyperparameter Settings}\label{appendix:hyperparam-setting}
\begin{table}[h]
\centering
\caption{Hyper-parameters for Each Dataset in Phase 1~(P1) and Phase 2~(P2)}
\resizebox{\columnwidth}{!}{
\begin{tabular}{cccccc}
  \hline
  Dataset & lr~(P1)  & dropout~(P1) & $\gamma$ & lr~(P2) & dropout~(P2) \\
  \hline
  Amazon-photo  & 0.005  & 0.2 & 55 & 0.001 & 0.6  \\
  Amazon-computer & 0.0005 & 0.2 & 15 & 0.001 & 0.4 \\
  Coauthor-cs & 0.001 & 0.2 & 55 & 0.01 & 0.6 \\
  Coauthor-physics & 0.01 & 0.2 & 15 & 0.01 & 0.6 \\
  Wiki-cs & 0.0005 & 0.2 & 15 & 0.001 & 0.4 \\
  \hline
\end{tabular}
}
\label{table:Parameters}
\end{table}

\name~has two training phases (Section~\ref{sec:Optimization}), and we conduct parameter search for each phase on each dataset. In both phases, we conduct separate grid searches 
for the learning rate (lr) within {0.01, 0.001, 0.0005} and the dropout rate within {0.2, 0.4, 0.6}, with different optimal values selected for each phase based on the validation set. Additionally, in the first phase, we also perform a grid search for $\gamma$ within {15, 55, 95, 135}. Table~\ref{table:Parameters} lists the parameter settings that achieve the optimal performance on each dataset. 

\section{Additional Experiment Results}\label{sec:additional result}

Here we present the results with AUROC and AUPR for both misclassification and OOD detection using models selected by validation accuracy on Amazon-photo dataset in Table~\ref{tab:result2}.  As we can see, the results align with our main findings and support the robustness of our method. However, we note that AUROC and AUPR can be misleading for misclassification detection, as each model defines positives and negatives differently~\cite{chen2024redl}. These metrics reflect internal ranking quality rather than true prediction reliability. Thus, we use AURC as the primary metric for the misclassification detection task.

\section{Scalability Analysis}\label{sec:scalability analysis}

\begin{figure}[h]
\centering
\includegraphics[width=1\columnwidth]{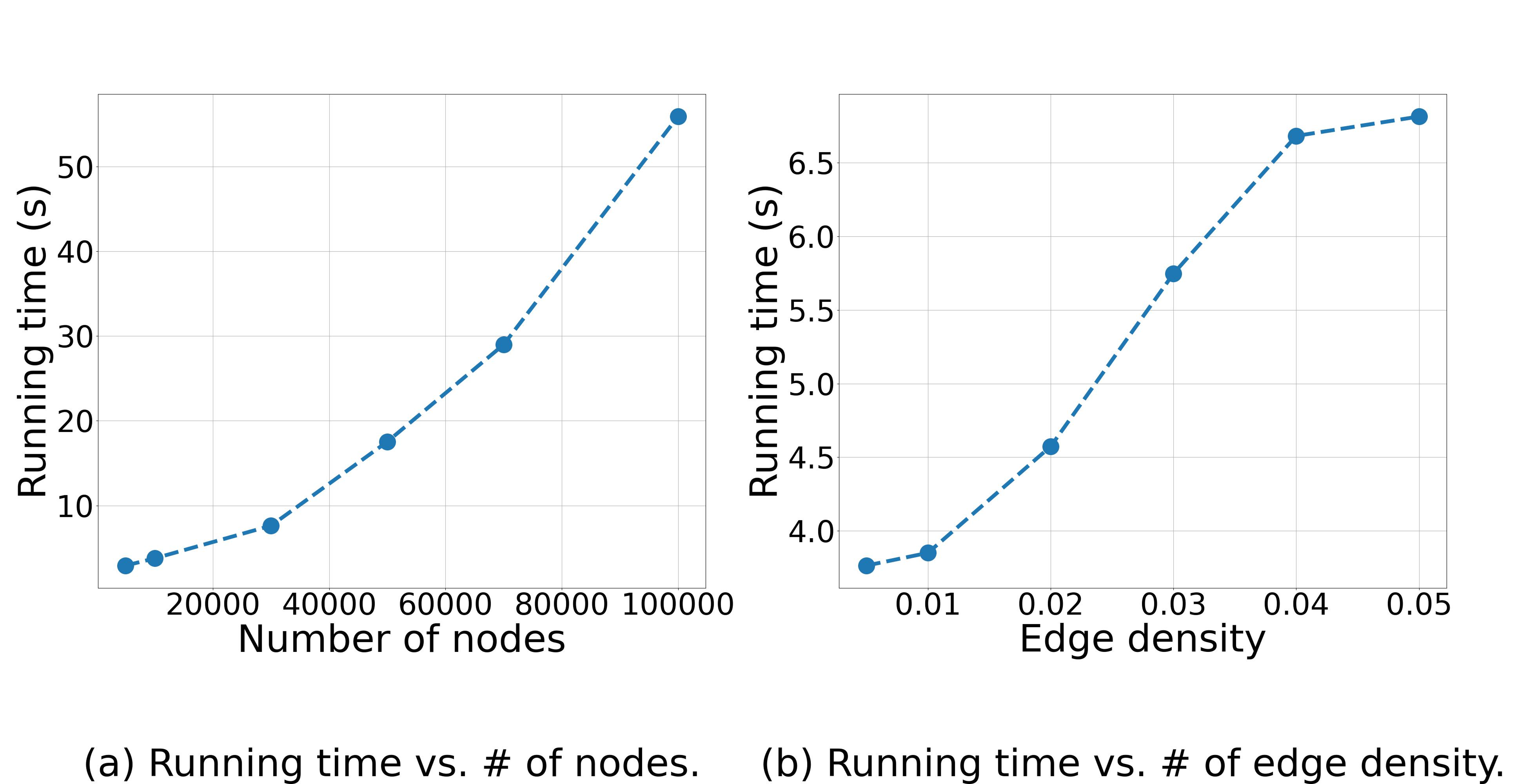} 
\caption{Running time vs. the number of nodes and edge density.}
\Description{figure description}
\label{fig:scalability}
\end{figure}

\begin{table}[h]
\centering
\caption{Comparison of Training Time, Inference Time, and Memory Usage on ogbn-arxiv }
\resizebox{\columnwidth}{!}{
\begin{tabular}{c|c|c|c}
    \hline
    Method  & \makecell{Training Time\\(s)/epoch} & \makecell{Inference Time\\(s)/epoch} & \makecell{Memory Usage\\(MB)} \\
    \hline
    ODIN     & 0.020  & 2.66 & 862  \\
    MaxLogit & 0.019  & 2.4  & 862  \\
    GNNSafe  & 0.032  & 2.46 & 1095 \\
    CRL      & 0.020  & 2.5  & 863  \\
    DOCTOR   & 0.019  & 2.5  & 862  \\
    GKDE     & 21.17  & 2.4  & 878  \\
    GPN      & /      & /    & /    \\
    Ours     & 5.05   & 2.5  & 1400 \\
    \hline
\end{tabular}
}
\label{table:complexity}
\end{table}

Here we analyze the scalability of \name, by reporting the running time of \name~on a series of synthetic graphs with increasing sizes (i.e., the number of nodes and the edge density), we generate the synthetic graphs via the ER algorithm \cite{erdds1959random}. In Figure \ref{fig:scalability}(a), we fix the edge density to be 0.005 and gradually increase the number of nodes from 5000 to 100000. In Figure \ref{fig:scalability}(b), we fix the number of nodes to 10000 and increase the edge density from 0.005 to 0.05. 
From Figure \ref{fig:scalability}, we observe that the running time of \name~scales efficiently with both the number of nodes and edge density. As the number of nodes increases, the running time shows steady growth, with a slight superlinear increase at larger scales. This might be due to the increasing number of logical operations needed for Dissonance Reasoning and Vacuity Reasoning. While regarding the scalability with respect to edge density, even though the edge density increases by 10x, the running time only increases less than 2x, demonstrating the reasonable computational efficiency of \name.

What's more, we also evaluated \name's scalability on ogbn-arxiv with 169,343 nodes and 1,166,243 edges, which is significantly larger than our main experimental datasets like Amazon-computer (13,752 nodes). The results for \name~and all baseline models are shown in Table~\ref{table:complexity}. Simple baseline models like ODIN and DOCTOR show minimal overhead with ~0.02s training time and ~860MB memory usage. EviNet requires higher resources (5.05s training time, 1400MB memory) due to its Beta embedding and context integration components. Notably, while GKDE, which is also a subjective logic-based method, uses less memory but requires much longer training time (21.17s per epoch), GPN was unable to process this dataset due to out-of-memory issues. While \name~has higher computational costs than simpler baselines, this trade-off enables its superior performance in both misclassification and OOD detection tasks.

\section{Computational Complexity Compared with Regular Graph Embedding}
\label{sec:complexity_comparison}

We analyze the computational complexity of our proposed model in comparison to standard graph embedding approaches. When the Beta embedding and reasoning modules are removed from \name, the remaining architecture reduces to a standard evidential network with a GNN encoder. The complexity of this baseline is given by $\mathcal{O}(mH + nH^2 + nKH)$, where $m$ is the number of edges, $n$ is the number of nodes, $H$ is the hidden dimension, and $K$ is the number of classes.

\name~extends this baseline by incorporating class-aware reasoning through Beta embeddings. This includes class-level disjunction reasoning, context-sensitive class-specific GCNs, and the estimation of dissonance and vacuity. These components contribute an additional complexity of $\mathcal{O}(K(mH + nH^2) + nK^2 + KnD)$, where $D$ represents the dimension used in reasoning. The overall complexity thus grows with the number of classes $K$.

While this introduces additional computational cost, it enables \name~to explicitly model both dissonance and vacuity uncertainty, resulting in substantially improved performance on misclassification detection and out-of-distribution detection. These capabilities are beyond the reach of standard graph embedding methods or GNNs without explicit reasoning mechanisms.

\end{document}